\documentclass[10pt,twocolumn,letterpaper]{article}

\usepackage{cvpr}
\usepackage{times}
\usepackage{epsfig}
\usepackage{graphicx}
\usepackage{amsmath}
\usepackage{amssymb}

\usepackage[ruled,linesnumbered]{algorithm2e}
\usepackage{glossaries}
\makeglossaries
\usepackage{amsthm}
\usepackage{breqn}
\usepackage{xcolor}
\usepackage{subfigure}
\usepackage{comment}
\usepackage{pgfplots}
\usetikzlibrary{patterns}
\usepackage{tkz-graph}

\renewcommand{\vec}[1]{\ensuremath{\mathbf{#1}}}

\newcommand{\mat}[1]{\ensuremath{\mathbf{#1}}}

\newcommand{\norm}[1]{\left\lVert#1\right\rVert}
\newcommand{\T}{\textnormal{T}}

\DeclareMathOperator*{\argmin}{arg\,min}
\newtheorem{experiment}{Experiment}
\definecolor{color0}{RGB}{255, 255, 255}
\definecolor{color1}{RGB}{230, 245, 255}
\definecolor{color2}{RGB}{179, 224, 255}
\definecolor{color3}{RGB}{102, 194, 255}
\definecolor{color4}{RGB}{26, 163, 255}
\definecolor{color5}{RGB}{0, 122, 204}
\newacronym{svd}{SVD}{singular value decomposition}
\newacronym{lr}{LR}{logistic regression}
\newacronym{llr}{LLR}{linear logistic regression}
\newacronym{ml}{ML}{machine learning}
\newacronym{blr}{BLR}{bilinear logistic regression}
\newacronym{sr}{SR}{soft-max regression}
\newacronym{bsr}{BSR}{bilinear soft-max regression}
\newacronym{mnist}{MNIST}{MNIST dataset}
\newacronym{svm}{SVM}{support vector machines}

\usepackage[pagebackref=true,breaklinks=true,letterpaper=true,colorlinks,bookmarks=false]{hyperref}

\cvprfinalcopy 

\def\cvprPaperID{****} 

\begin{document}

\title{Bilinear Models for Machine Learning}

\author{Tayssir Doghri, Leszek Szczecinski, Jacob Benesty, and Amar Mitiche\\
Institut National de la Recherche Scientifique\\
{\tt\small \{tayssir.doghri,leszek,benesty,mitiche\}@emt.inrs.ca}
}

\maketitle

\begin{abstract}
In this work we define and analyze the bilinear models which replace the conventional linear operation used in many building blocks of \gls{ml}. The main idea is to devise the \gls{ml} algorithms which are adapted to the objects they treat. In the case of monochromatic images, we show that the bilinear operation exploits better the structure of the image than the conventional linear operation which ignores the spatial relationship between the pixels. This translates into significantly smaller number of parameters required to yield the same performance. We show numerical examples of classification in the MNIST data set.
\end{abstract}

\section{Introduction}
The purpose of this work is to define and analyze the bilinear model for the use in \gls{ml}, as well as, to propose the suitable learning algorithms. We focus on the simplest \gls{ml} model defined through \gls{lr} composed of linear processing followed by a nonlinear activation function. Since the latter is a building block of many more advanced \gls{ml} models such as neural networks, the first step is to understand the properties and learning algorithms in case of bilinear processing which replaces the linear one.

Our work is motivated by the fact that the typical \gls{ml} tasks such as classification often use data which, when originally acquired, has strong structural dependence between its elements. In particular, monochromatic images acquisition yields the structures which are naturally represented as matrices and there is often similarity/relationship between the pixels which are close to each others. 

A common approach to deal with any data in \gls{ml} is vectorization which lists data elements in a predefined order (\eg row by row). The obvious advantage is that the resulting vectors can be treated by generic \gls{ml} algorithms such as those described in \cite{bishop2006}. 

On the other hand, the loss of structure is intuitively counterproductive and here we want to explore the possibility of using operations which are defined taking the data structure into account, \eg the relationship between neighbour pixels. To this end, we propose to replace the linear operation, which weights and sums all the pixels in the image, with a sum of bilinear operators.

The advantage of such \gls{blr} is that, by exploiting the input data in its original structure, we may directly access the useful information and thus define the  model using less parameters. More specifically, for an input image represented as a matrix $\mat{X} \in \mathbb{R}^{M \times N}$, the conventional linear operation first creates the vector $\vec{x}\in \mathbb{R}^{MN}$ which requires $MN$ parameters of linear combiner, while a bilinear form requires $M + N$ parameters; the difference which becomes important for large $M$ and $N$. When using the sum of $L$ bilinear forms, the number of coefficients grows linearly with $L$. We show that exploiting the spatial structure of the image reduces significantly the number of coefficients required by the classifier, which also points to the overparametrization problem inherent in \gls{ml} which is blind to the image structure.

\subsection{Contribution and related work}
\label{section: related work}

The idea of bilinear structures to replace the generic linear vector-based processing is not entirely new. For example, it was already adopted in the context of acoustic signal processing to identify the response of the accoustic channel \cite{jacob2017,jacob2019}.

In the context of \gls{ml}, the bilinear forms were also used to replace the linear processing in \gls{svm} \cite{NIPS2009_3789}, and to make classification based on \gls{lr} for multi-channel medical data \cite{eegArticle}. Similarly \cite{tensor-based2017} proposed a multilinear operation to deal with the multidimensional data represented by a tensor.

The main difference with the previous work is that, \cite{eegArticle} limited the considerations to rank-1 \gls{blr} and and \cite{tensor-based2017} use rank-1 multilinear form; here, we propose high-rank \gls{blr}; this is possible thanks to the algorithms we devise that rely on the alternate optimization. This stands in contrast to the approach adopted by \cite{eegArticle} which defined the global optimization problem, which does not guarantee the convergence.

Our work is closest in the spirit to \cite{jacob2017,jacob2019} which were mainly concerned with tracking of time-varying models, while we deal with static data for classification which allows us to devise new efficient alternate optimization algorithms for high-rank \gls{blr}.

\subsection{Structure}

This paper is organized as follows: Section~\ref{section: bilinear logistic regression} introduces the concept of bilinear model applied to logistic regression along with its interpretation, while  the learning algorithm is shown in Sec.~\ref{Sec:Training}, where the regularization is also discussed. Section~\ref{section: Generalisation to multiclass classification problem} generalizes softmax regression using bilinear forms to treat multiclass classification problems. The experimental results are presented as examples to illustrate the behaviour of the proposed models. We conclude the work in Sec.~\ref{Sec:Discussion}.

\section{Bilinear logistic regression}
\label{section: bilinear logistic regression}

Before talking about bilinear logistic regression (\gls{blr}), which is the focus of our work, it is convenient to rediscuss the conventional, \gls{llr}.

\subsection{Conventional logistic regression}\label{Sec:LR}
The problem is defined as follows: from the observed features gathered in the vectors $\vec{x}_t$, indexed by $t$, we want to obtain the estimate of the posterior probability of the classes $C_t\in\{0,1\}$ to which $\vec{x}_t$ belongs; that is, we want to find $y_t=\Pr\{C_t=1|\vec{x}_t\}$.\footnote{In this binary classification case, we have $\Pr\{C_t=0|\vec{x}_t\}=1-y_t$.}

\gls{llr} refers to a model which approximates $y_t$ using a non-linear function applied to a linear transformation of $\vec{x}_t$ \cite{bishop2006}:
\begin{align}
\label{model: LR}
    y_t &= f\big( z_t \big),\\
\label{z.w}
    z_t&=\vec{w}^\T\vec{x}_t,
\end{align}
where $\vec{w}$ contains the weights and $f(\cdot)$ is the \textit{logistic function} defined as
\begin{align}
    f(z)=\frac{1}{1+\textnormal{e}^{-z}}.
\end{align} 

Then, given the training data $\{(\vec{x}_t,c_t)\}_{t=1}^T$, where $c_t\in\{0,1\}$ is the class of the vector $\vec{x}_t$ and $T$ is the number of training examples, we want to find the most appropriate weights $\vec{w}$. This \emph{learning} is done via optimization:
\begin{align}
\label{eq:optimization}
    \hat{\vec{w}}&=\argmin_{\vec{w}} J(\vec{w}),  \\
    \label{J.w}
    J(\vec{w}) &= V(\vec{w}) +  \alpha R(\vec{w}),
\end{align}
where 
\begin{align}\label{eq:crossentropy}
  V(\vec{w})&=-\frac{1}{T}\sum_{t=1}^T \Big[ c_t \log y_t + (1-c_t)\log (1-y_t)\Big],
\end{align}
is the cross-entropy (or, the negated likelihood of the classes) which ensures adequacy of the model fit to the data, $\alpha$ is the regularization parameter, and $R(\vec{w})$ is the regularization function, often the squared norm of $\vec{w}$, \ie,
\begin{align}\label{R.w}
    R(\vec{w})&=\frac{1}{2}\|\vec{w}\|_2^2.
\end{align}

Since $J(\vec{w})$ is convex, the solution of \eqref{eq:optimization} is unique and can be sought using the gradient which is calculated as
\begin{align}
\label{gradient:w}
    \nabla_{\vec{w}} J(\vec{w}) = \frac{1}{T} \sum_{t=1}^{T} (y_t - c_t) \vec{x}_t
    + \alpha \vec{w}.
\end{align}

In this most common approach to the \gls{lr}, the features are represented as a vector $\vec{x}_t$ to simplify the calculations as shown above; any explicit relationship between the elements of $\vec{x}_t$ is deliberately ignored. In particular, and this is the focus of this work, if the features $\vec{x}_t$ are originally represented by a matrix $\mat{X}_t \in \mathbb{R}^{M \times N}$, which occurs naturally when $\mat{X}_t$ is a monochromatic image, the spatial information between the pixels in $\mat{X}_t$ is lost after vectorization of $\mat{X}_t$ into $\vec{x}_t$.

The vectorization not only removes the structure but makes the interpretation of the results less natural. In fact, it is much more convenient to represent the relationship \eqref{z.w} using a matrix notation
\begin{align}
    z_t&=\langle \vec{W}, \vec{X}_t \rangle,\label{LR:matrix}
\end{align}
where $\langle \cdot, \cdot \rangle$ is the inner product of its arguments; $\mat{W}$ is a matricized version of $\vec{w}$ in \eqref{z.w} and may be seen as the spatial filter applied to the image represented by $\mat{X}_t$.

\subsection{Bilinear logistic regression}\label{Sec:BLR}
While the matrix-based calculation of $z_t$ in \eqref{LR:matrix} provides the interpretation of the weights $\mat{W}$ in the domain of images, it is done merely by reorganization of the elements. 

On the other hand, the relationship between the pixels is not accounted for. To address this issue we propose the bilinear processing defined as follows:
\begin{equation}
    z_t = \sum_{l=1}^{L} \vec{a}_l^\T \mat{X}_t \vec{b}_l,
    \label{model: blr}
\end{equation}
where the left- and right-hand side elements of the  bilinear transformation, $\vec{a}_l$ and $\vec{b}_l$, are gathered in matrices
\begin{align}
\mat{A}&=[\vec{a}_1,\ldots,\vec{a}_L]\\
\mat{B}&=[\vec{b}_1,\ldots,\vec{b}_L].
\end{align}

Indeed, we can now interpret the vectors $\vec{a}_l$ and $\vec{b}_l$ as filters acting, respectively, on the columns and rows of the image $\mat{X}$. 

Two observations are in order regarding the proposed \gls{blr}:
\begin{itemize}
    \item For $L=\min\{M,N\}$, \gls{blr} defined in \eqref{model: blr} is equivalent to the linear logistic regression \eqref{model: LR} under suitable choice of $\mat{A}$ and $\mat{B}$; we demonstrate it in Sec.~\ref{Sec:equivalence}.
    \item For $L<\min\{M,N\}$, \gls{blr} introduces correlation between the pixels in the same columns and the same rows; this is shown in Sec.~\ref{Sec:Interpretation}.
\end{itemize}

\subsection{Equivalence between \gls{llr} and \gls{blr}}\label{Sec:equivalence}

To demonstrate the equivalence between \eqref{model: blr} and \eqref{LR:matrix}, we rewrite the latter as
\begin{equation}\label{log.regression.matrix}
    z =  \langle \mat{W}, \mat{X} \rangle=\textnormal{Tr}(\mat{W}^\T \mat{X}),
\end{equation}
where $\textnormal{Tr}(\cdot)$ denotes the trace of a matrix, and temporarily we removed the subindexing with $t$.

The matrix $\mat{W}$ may be decomposed using \gls{svd} as follows:
\begin{equation}
    \label{SVD.decomposition}
    \mat{W} = \mat{U}\mat{S}\mat{V}^\T =\sum_{l=1}^L s_l \vec{u}_l \vec{v}_l^\T,
\end{equation} 
where $\mat{U}=[\vec{u}_1, \cdots, \vec{u}_M] \in \mathbb{R}^{M\times M}$ and $\mat{V}=[\vec{v}_1, \cdots, \vec{v}_N] \in \mathbb{R}^{N\times N}$ are orthogonal matrices and $\mat{S}=\textnormal{diag}(\mat{S}_1, \textbf{0})$, $\mat{S}_1 = \textnormal{diag}(s_1,\cdots,s_L)$. 

Thus, we get
\begin{align}
z
     &= \textnormal{Tr}\Big[ \big(\sum_{l=1}^L s_l \vec{u}_l \vec{v}_l^\T \big)^\T \mat{X} \Big] \\
     &= \sum_{l=1}^L s_l \vec{u}_l^\T \mat{X} \vec{v}_l.
\end{align}
By setting $\vec{a}_l = \sqrt{s_l} \vec{u}_l$ and $\vec{b}_l = \sqrt{s_l} \vec{v}_l$, we obtain \eqref{model: blr}; we can also rewrite \eqref{SVD.decomposition} as
\begin{align}
\label{W.AB}
    \mat{W}&=\sum_{l=1}^L \vec{a}_l \vec{b}_l^\T.
\end{align}

\subsection{Induced conditional dependence}\label{Sec:Interpretation}
The useful insight into the bilinear model we propose may be obtained looking at the implicit generative model underlying the classification principle of the linear logistic regression. 

Namely, if we assume that the distribution of the features $\vec{x}$ conditioned on the class $C$ is given by 
\begin{align}\label{generative.model}
    p(\vec{x}|C=1)\propto\exp(\vec{w}^\T\vec{x})g(\vec{x}),
\end{align}
where $g(\vec{x})$ is an arbitrary function independent of the class $c$, \eqref{model: LR} follows, similarly as in \cite[Ch.~4.2.1]{bishop2006}. This relationship also means that, conditioned on the class $C$, we know the weight $w_i$ and thus the features $x_i$ are independent. This is equivalent to assuming that \eqref{model: LR} implements a naive Bayes rule  \cite[Ch.~6.6.3]{Hastie_book}.

This can be seen in the graphical representation of the probabilistic dependencies shown in Fig.~\ref{figure:dependecies}, where we use the formalism of representing the dependence in Bayesian network via arrows connecting the parent (arrow's tail) and the child (arrow's head) \cite[Ch.~3.3]{Barber12_Book}. This also corresponds to the conditional probability. 

Then, knowing $C$, ``blocks'' any path connecting the weights $w_{i,j}$ which are thus (conditionally) independent. 

On the other hand, in the case of the \gls{blr} with $L=1$ we can rewrite \eqref{W.AB} as $\mat{W}=\vec{a}\vec{b}^\T$, \ie each term of the matrix $\mat{W}$ can be written as $w_{i,j}=a_i b_j$, where $a_i$ and $b_j$ are elements of the vectors $\vec{a}$ and $\vec{b}$. This relationships is shown in Fig.~\ref{figure:dependecies}b and we  see that knowing $C$ does not block the paths between the weights $w_{i,j}$ and they remain connected through the elements of $a_i$ and $b_j$. For example, there exist a path connecting $w_{1,1}$ and $w_{1,N}$ (via variable $a_1$) and which does not include $C$. We hasten to say that this merely says that the elements $w_{i,j}$ are not structurally independent, their independence can still be obtained with the appropriate choice of the values in the vectors $\vec{a}$ and $\vec{b}$.

We emphasize also that we do not need the generative models \eqref{generative.model} to perform the classification. We rather use it in Fig.~\ref{figure:dependecies} to clarify the difference between the conventional \gls{llr} and the \gls{blr}. The most important conclusions is that while, the dependencies between the features (here, the pixels) are often imposed by a non-linear transformation of $\vec{x}$ (such as, \eg squaring, see \cite[Ch.~4.1]{Hastie_book}), here they are imposed by the hierarchical structure of the bilinear operation.

\begin{figure}
	\centering
	\begin{subfigure}[]
		\centering
    		\begin{tikzpicture}[font=\footnotesize]
    		\tikzset{VertexStyle/.style = {%
                    shape = circle,
                    minimum size = 27pt,draw}}
                \SetVertexMath                %
               
               \Vertex[x=-0.5,y=3.5]{x_{1,1}}
               \Vertex[x=-0.5,y=2.3]{x_{1,2}}
               \Vertex[x=-0.5,y=0.5]{x_{M,N}}
               \Vertex[x=1.5,y=3.5]{w_{1,1}}
               \Vertex[x=1.5,y=2.3]{w_{1,2}}
               \Vertex[x=1.5,y=0.5]{w_{M,N}}
               \Vertex[x=3.5,y=2]{C}
               
               \GraphInit[vstyle=Empty]
               \Vertex[x=-0.5,y=1.4,L=\vdots]{p3}
               \Vertex[x=1.5,y=1.4,L=\vdots]{p33}
               %
               \Edge[style={->}](w_{1,1})(x_{1,1})
               \Edge[style={->}](w_{1,2})(x_{1,2})
               \Edge[style={->}](w_{M,N})(x_{M,N})
               \Edge[style={->}](C)(w_{M,N})
               \Edge[style={->}](C)(w_{1,1})
               \Edge[style={->}](C)(w_{1,2})
            \end{tikzpicture}
	\end{subfigure}
	\vfill
	\begin{subfigure}[]
		\centering
    		\begin{tikzpicture}[font=\footnotesize]
    		\tikzset{VertexStyle/.style = {%
                    shape = circle,
                    minimum size = 27pt,draw}}
                \SetVertexMath
                %
               \Vertex[x=0,y=3.9]{x_{1,1}}
               \Vertex[x=0,y=2.8]{x_{1,2}}
               \Vertex[x=0,y=1]{x_{1,N}}
               \Vertex[x=1.7,y=3.9]{w_{1,1}}
               \Vertex[x=1.7,y=2.8]{w_{1,2}}
               \Vertex[x=1.7,y=1]{w_{1,N}}
               \Vertex[x=0,y=-1.5]{x_{M,1}}
               \Vertex[x=0,y=-2.6]{x_{M,2}}
               \Vertex[x=0,y=-4.5]{x_{M,N}}
               \Vertex[x=1.7,y=-1.5]{w_{M,1}}
               \Vertex[x=1.7,y=-2.6]{w_{M,2}}
               \Vertex[x=1.7,y=-4.5]{w_{M,N}}
               \Vertex[x=6.5,y=0]{C}
               \Vertex[x=4,y=4]{a_{1}}
               \Vertex[x=4,y=-4]{a_{M}}
               \Vertex[x=4,y=1.6]{b_{1}}
               \Vertex[x=4,y=0.3]{b_{2}}
               \Vertex[x=4,y=-1.8]{b_{N}}
               
               \GraphInit[vstyle=Empty]
               \Vertex[x=0,y=1.95,L=\vdots]{p3}
               \Vertex[x=1.7,y=1.95,L=\vdots]{p33}
               \Vertex[x=0,y=-3.46,L=\vdots]{p4}
               \Vertex[x=1.7,y=-3.46,L=\vdots]{p44}
               \Vertex[x=0.85,y=-0.2,L=\vdots]{p5}
               %
               \Edge[style={<-}](x_{1,1})(w_{1,1})
               \Edge[style={<-}](x_{1,2})(w_{1,2})
               \Edge[style={<-}](x_{1,N})(w_{1,N})
               \Edge[style={<-}](w_{1,1})(b_{1})
               \Edge[style={<-}](w_{1,2})(b_{2})
               \Edge[style={<-}](w_{1,N})(b_{N})
               \Edge[style={<-}](w_{1,1})(a_{1})
               \Edge[style={<-}](w_{1,2})(a_{1})
               \Edge[style={<-}](w_{1,N})(a_{1})
               \Edge[style={<-}](x_{M,1})(w_{M,1})
               \Edge[style={<-}](x_{M,2})(w_{M,2})
               \Edge[style={<-}](x_{M,N})(w_{M,N})
               \Edge[style={<-}](w_{M,1})(b_{1})
               \Edge[style={<-}](w_{M,2})(b_{2})
               \Edge[style={<-}](w_{M,N})(b_{N})
               \Edge[style={<-}](w_{M,1})(a_{M})
               \Edge[style={<-}](w_{M,2})(a_{M})
               \Edge[style={<-}](w_{M,N})(a_{M})
               \Edge[style={<-}](b_{1})(C)
               \Edge[style={<-}](b_{2})(C)
               \Edge[style={<-}](b_{N})(C)
               \Edge[style={<-}](a_{1})(C)
               \Edge[style={<-}](a_{M})(C)

            \end{tikzpicture}
	\end{subfigure}
    \caption{The implicit generative model behind a)~conventional logistic regression, and b)~bilinear logistic regression for $L=1$.}
	\label{figure:dependecies}
\end{figure}

\section{Model training}\label{Sec:Training}

Our objective is to learn the weights vectors $\vec{a}_l$ and $\vec{b}_l$ gathered in matrices $\mat{A}$ and $\mat{B}$ directly from the training set $\{(\mat{X}_t, c_t)\}_{t=1}^{T}$.

As in the conventional \gls{llr}, it will done by optimization 
\begin{align}
\label{optim.A.B}
    [\hat{\mat{A}},\hat{\mat{B}}]=\argmin_{\mat{A},\mat{B}} J(\mat{A}, \mat{B}),\\
\label{J.A.B}    
    J(\mat{A}, \mat{B}) = V(\mat{A}, \mat{B}) + \alpha R(\mat{A}, \mat{B}),
\end{align}
where $V(\mat{A}, \mat{B})$ is the bilinear version of the cross-entropy defined in \eqref{eq:crossentropy} and the regularization term $R(\mat{A}, \mat{B})$ plays the same role as $R(\vec{w})$ in \eqref{J.w}. While it is not immediately obvious how to choose this function, for the purpose of the discussion about training we assume it takes form similar to \eqref{R.w}, namely
\begin{align}
\label{R.AB.sum} 
    R(\mat{A}, \mat{B}) = \frac{1}{2}\sum_{l=1}^L \big( \norm{\vec{a}_l}_{2}^2 + \norm{\vec{b}_l}_{2}^2 \big).
\end{align}

We note that i)~the function \eqref{J.w}, $J(\vec{w})$, was convex in $\vec{w}$ due to the linear relationship between $z$ and $\vec{w}$ see \eqref{z.w}, and ii)~$z$ in \eqref{model: blr} is \emph{not} linear in $\vec{a}_l$ and $\vec{b}_l$. Thus, the convexity of $J(\mat{A},\mat{B})$  with respect to $\vec{a}_1,\ldots,\vec{a}_L,\vec{b}_1,\ldots,\vec{b}_L $ is not guaranteed so the direct use of the gradient methods should be discouraged.

On the other hand, we note that fixing all, but one, terms in $\mat{A}$ and $\mat{B}$, the optimization problems is transformed into the one we already dealt with in the case of \gls{llr} in Sec.~\ref{Sec:LR}.

This can be easily seen taking the gradient of the cost function with respect to $\vec{a}_l$ and $\vec{b}_l$:
\begin{align}
\label{cost gradient: blr with repect to (R)}
    \nabla_{\vec{a}_l} J(\mat{A}, \mat{B}) &= \frac{1}{T} \sum_{t=1}^{T} (y_t - c_t) \mat{X}_t \vec{b}_l + \alpha \vec{a}_l,
\\
\label{cost gradient: blr with repect to (L)}
    \nabla_{\vec{b}_l} J(\mat{A}, \mat{B}) &= \frac{1}{T} \sum_{t=1}^{T}(y_t - c_t) \vec{a}_l^\T \mat{X}_t + \alpha \vec{b}_l,
\end{align}
which yields the equations similar to the one we show in \eqref{gradient:w}.

Thus, the function $J(\mat{A}, \mat{B})$ is convex with respect to $\vec{a}_l$ or $\vec{b}_l$ if all other vectors are fixed. This suggests the use of alternate optimization procedure, where we optimize with respect to the vectors $\vec{a}_l$ or $\vec{b}_l$ one at the time, as described in Algorithm~\ref{Algo:BLR}. Since each of the problems we solve is convex we can use the gradient-based methods as we did before; that is, we optimize with respect to $\vec{a}_l$ (lines \ref{algo: begin al}-\ref{algo: end al} in Algorithm~\ref{Algo:BLR}) then with respect to $\vec{b}_l$ (lines \ref{algo: begin bl}-\ref{algo: end bl} in Algorithm~\ref{Algo:BLR}).

The optimization is done in multiple steps denoted by $i=1,\ldots, i_{\textnormal{max}}$. The initialization of the weights (for $i=1$) is important to speed-up the convergence. 

First of all, for a given $l$, we want to learn the weights $\vec{a}_l$ and $\vec{b}_l$ knowing $\vec{a}_k, \vec{b}_k, k<l$ but assuming the null contribution from the weights $\vec{a}_k, \vec{b}_k, k>l$. In other words we treat each rank $l$ as providing additional approximation level. This explains why $\vec{a}_l$ is initialized to zero. It is of course possible to initialize randomly all weights $\vec{a}_k$ and $\vec{b}_k$ but this slows down the convergence because training of $\vec{a}_l$ and $\vec{b}_l$ we are affected by the random values attributed to $\vec{a}_k, \vec{b}_k, k>l$.

Second, for a given rank $l$, since the vectors $\vec{a}_l$ and $\vec{b}_l$ affect $z_t$ through multiplication, we cannot set them both to zero, as this would produce zero gradient, see \eqref{cost gradient: blr with repect to (R)} and \eqref{cost gradient: blr with repect to (L)}; this explains the random initialization of $\vec{b}_l$ (line \ref{algo: random initblr} of Algorithm~\ref{Algo:BLR}). We also noted that the initial orthogonalization of the vectors $\vec{b}_l, l=1, \ldots, L$  (lines \ref{algo: begin orht proj}-\ref{algo: end orht proj} of Algorithm~\ref{Algo:BLR}) improves the convergence rate. This approach is inspired by the \gls{svd} decomposition \eqref{SVD.decomposition}. We note however, that this is not a formal constraint on the solution. In fact, imposing such a constraint slightly deteriorates the performance.

The last comment to be made concerns the non-uniqueness of the solution which is due to the very structure \eqref{W.AB}, from which it is clear that any solution in the form $\big(\beta \vec{a}_l, \frac{1}{\beta} \vec{b}_l\big)$ yields exactly the same results, because the product in \eqref{W.AB} cancels out any $\beta\neq 0$.

\begin{algorithm}\label{Algo:BLR}
    \SetAlgoLined
    \caption{Training of \gls{blr}}
    \label{algo: minimize cost: blr}
    
    \text{Initialization:}\\
    $\vec{a}_l = \vec{0}, l=1,\ldots, L$\\
    $\vec{b}_l = \text{drawn from a uniform distribution over [-1,1]}$\\\label{algo: random initblr}
    \For{$l = 2, \cdots,  L$}{
    \label{algo: begin orht proj}
        $\mat{S}_{l-1} = \Big[ \frac{\vec{b}_1}{\norm{\vec{b}_1}}, \cdots, \frac{\vec{b}_{l-1}}{\norm{\vec{b}_{l-1}} }\Big] $  
          
        $\vec{b}_l \leftarrow \vec{b}_l - \mat{S}_{l-1}\big(\mat{S}_{l-1}^\T \vec{b}_l\big)$
    
    }
    \label{algo: end orht proj}
    
    \text{Optimization:}\\
    \For{$i=1, \cdots,  i_{\textnormal{max}}$}{
        \For{$l = 1, \cdots, L$}{
            \While{$\vec{a}_l$ \textnormal{not converged} }{
                \label{algo: begin al}
                $ \vec{g}_l = \nabla_{\vec{a}_l} J(\mat{A}, \mat{B})$
                
                $\mat{D}_{l} = [ \vec{0}, \cdots, \vec{g}_l, \cdots, \vec{0} ]$ \label{algo: gradl}
                
                $\hat{\eta} \approx \argmin_{\eta}{J(\mat{A}-\eta\mat{D}_{l}, \mat{B})}$
                
                $\vec{a}_l \leftarrow \vec{a}_l - \hat{\eta} \vec{g}_l$
                
            }
            \label{algo: end al}
            \While{$\vec{b}_l$ \textnormal{not converged}}{
                \label{algo: begin bl}
            	$ \vec{g}_l = \nabla_{\vec{b}_l} J(\mat{A}, \mat{B})$
            
                $\mat{D}_{l} = [ \vec{0}, \cdots, \vec{g}_l, \cdots, \vec{0} ]$ \label{algo: gradr}
            	
            	$\hat{\eta} \approx \argmin_{\eta}{J(\mat{A}, \mat{B}-\eta\mat{D}_{l})}$
            	
            	$\vec{b}_l \leftarrow \vec{b}_l - \hat{\eta} \vec{g}_l$
            	
            }
            \label{algo: end bl}
        }
    }
\end{algorithm}

\subsection{Regularization strategies}\label{Sec:regularization}

The choice of the regularization function is often dictated by the simplicity of the resulting optimization procedure thus, the choice of \eqref{R.AB.sum} is justified by the simplicity of gradient calculation. 

Taking into account the fact that the bilinear filtering approach is equivalent to the linear counterpart, see Sec.~\ref{Sec:equivalence}, it might be interesting to use the regularization function $R(\mat{A}, \mat{B})$ which is equivalent to the original \gls{lr} problem \eqref{R.w}; that is, using \eqref{W.AB} we would define
\begin{align}\label{R.AB.sum.w}
    R(\mat{A}, \mat{B})=\frac{1}{2} \|\sum_{l=1}^{L} \vec{a}_l \vec{b}_l^\T \|^2_{\textnormal{F}}
\end{align}
via Frobenius norm.

However, such a definition would lead to a burden in the calculation of the gradient $\nabla_{\vec{a}_l}R(\mat{A}, \mat{B})$ and $\nabla_{\vec{b}_l}R(\mat{A}, \mat{B})$, we thus opt for \eqref{R.AB.sum} or for
\begin{equation}
\label{regularization: prod}
	R(\mat{A}, \mat{B}) = \frac{1}{2}\sum_{l=1}^L \norm{\vec{a}_l}_{2}^2 \norm{\vec{b}_l}_{2}^2,
\end{equation}
which is actually equivalent to \eqref{R.AB.sum.w} for $L=1$.

Also, for $L=1$, the sum-regularization \eqref{R.AB.sum} will yield the same solution as product-regularization \eqref{regularization: prod}.

This can be seen easily noting that, without any loss of generality, the solutions based on \eqref{regularization: prod} may be forced to satisfy $\norm{\vec{a}_1}_2=\norm{\vec{b}_1}_2$ (as said before Sec.~\ref{Sec:regularization}, the product of terms is what matters, and it may be kept constant while normalizing). The same can be said about the sum-regularization \eqref{R.AB.sum} which is obviously minimized for $\norm{\vec{a}_1}_2=\norm{\vec{b}_1}_2$. In other word, only the norm $\|\vec{a}_1\|_2=\|\vec{b}_1\|_2$ affects the solutions for the sum-regularization and the product-regularization.

On the other hand, increasing the rank $L$ such equivalence cannot be guaranteed.

\begin{experiment}\label{Experiment.1v1}
\label{section: experiments}
To test the proposed approach, we consider MNIST dataset consisting of $M\times N$ grayscale images of handwritten digits  going from $0$ to $9$ \cite{mnist} with the size $M=N=28$. Thus the \gls{lr} requires $MN=784$ weights to represent $\vec{w}$, while the \gls{blr} requires $L(M+N)=56 L$ weights to represent $\mat{A}$ and $\mat{B}$.

We used the training set with different number of elements $T\in\{32, 128,512,1024,4096, 8192\}$.  The final classification accuracy was obtained from the testing set with size $T_{\textnormal{test}}=2000$. 
The product-regularization \eqref{regularization: prod} was applied and the regularization parameter $\alpha$ was chosen using cross-validation on the validation set composed of $T_{\textnormal{val}}=2000$ elements.

The results of pairwise comparison of digits $8$ vs. $9$ are shown in Fig.~\eqref{figure: compare 89} and the comparison $5$ vs. $8$ is shown in Fig.~\eqref{figure: compare 58}. 

We can observe that i)~ for $L=1$, the \gls{blr} is consistently outperformed by the \gls{llr}; this is due to the equivalence of the regularization functions for $L=1$ and smaller number of parameters in \gls{blr}, and ii)~the gap in recognition accuracy is practically filled using $L=2$; thus, with $112$ coefficients required to represent $\mat{A}$ and $\mat{B}$ we obtain essentially the same performance as the conventional, \gls{llr} which requires approximately seven times more coefficients; this indicates that ignoring the structure of the image leads to the overparametrization of the solution. 

\pgfplotstableread[row sep=\\,col sep=&]{
images & lr  \\
32     & 93.7   \\
128    & 96.95   \\
512    & 97.36  \\
1024   & 97.86   \\
4096   & 98.17   \\
8192   & 98.52   \\
}\lr
\pgfplotstableread[row sep=\\,col sep=&]{
images  & rank-1-blr	& rank-2-blr    	& rank-3-blr    	& rank-4-blr   		\\
32      & 90 			& 91.77           	& 92.69      		& 91.77         	\\
128     & 94.56 		& 95.88           	& 96.70       		& 96.29          	\\
512     & 94.77 		& 96.29            	& 97.51      		& 97.46         	\\
1024    & 95.73 		& 97.05           	& 97.96       		& 97.91           	\\
4096    & 95.78 		& 97.61            	& 98.37      		& 98.68           	\\
8192    & 95.68 		& 97.36           	& 98.52        	& 98.57          	\\
}\blr

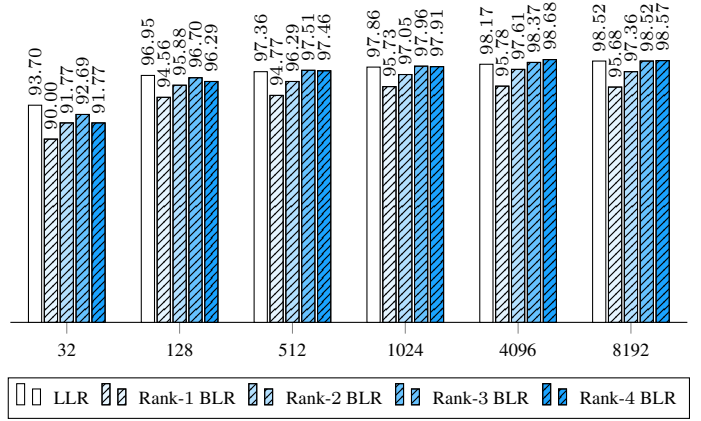
\begin{figure}
    \begin{tikzpicture}[font=\scriptsize]
    \begin{axis}[
       	ybar,
        bar width=5pt,
        x= 1.5cm,
        ybar=1pt,
        width=2\textwidth,
        height=.3\textwidth,
        hide y axis,
        axis x line*=bottom,
        legend style={at={(0.5,-0.2)}, anchor=north,legend columns=-1, column sep=0.1cm},
        symbolic x coords={32, 128, 512, 1024, 4096, 8192},
        xtick=data,
        nodes near coords,
        nodes near coords style={rotate=90, anchor=west, /pgf/number format/.cd,fixed zerofill,precision=2},
        ymin=70,ymax=100,
        ylabel={Accuracy},
    ]

    \addplot [style = {fill=color0}] table[x=images,y=lr]{\lr};	 
    \addplot [style = {fill=color1, mark=none, postaction={pattern=north east lines}}] table[x=images,y=rank-1-blr]{\blr};
    \addplot [style = {fill=color2, mark=none, postaction={pattern=north east lines}}] table[x=images,y=rank-2-blr]{\blr};
    \addplot [style = {fill=color3, mark=none, postaction={pattern=north east lines}}] table[x=images,y=rank-3-blr]{\blr};
    \addplot [style = {fill=color4, mark=none, postaction={pattern=north east lines}}] table[x=images,y=rank-4-blr]{\blr};
    
    \legend{LLR, Rank-$1$ BLR, Rank-$2$ BLR, Rank-$3$ BLR, Rank-$4$ BLR}
    \end{axis}
    \end{tikzpicture}
\caption{Comparaison of the accuracy of \gls{llr} and rank-$L$ \gls{blr} while performing classification of digits 8 and 9 with different training set size $T$ on \gls{mnist} data set.}
\label{figure: compare 89}
\end{figure}

\pgfplotstableread[row sep=\\,col sep=&]{
images & lr  \\
32     & 85.04   \\
128    & 90.31   \\
512    & 93.15  \\
1024   & 93.57   \\
4096   & 94.26   \\
8192   & 94.2   \\
}\lr
\pgfplotstableread[row sep=\\,col sep=&]{
images  & rank-1-blr	& rank-2-blr    	& rank-3-blr    	& rank-4-blr   		\\
32      & 72.72 			& 86.67           	& 86.04       		& 86.72         	\\
128     & 84.78 		& 89.83           	& 90.41       		& 90.62          	\\
512     & 86.3 		& 92.62           	& 92.62      		& 92.99         	\\
1024    & 86.88 		& 93.2           	& 94.1       		& 93.94          	\\
4096    & 87.73 		& 93.47            	& 94.31      		& 94.26          	\\
8192    & 87.94 		& 93.47           	& 94.41        	& 94.41          	\\
}\blr

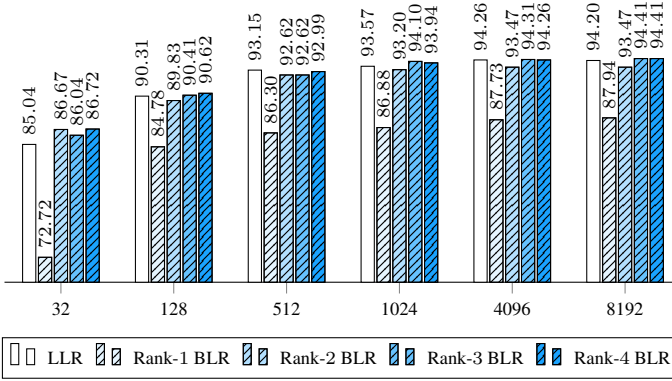
\begin{figure}
    \begin{tikzpicture}[font=\scriptsize]
    \begin{axis}[
       	ybar,
        bar width=5pt,
        x= 1.5cm,
        ybar=1pt,
        width=2\textwidth,
        height=.3\textwidth,
        hide y axis,
        axis x line*=bottom,
        legend style={at={(0.5,-0.2)}, anchor=north,legend columns=-1, column sep=0.1cm},
        symbolic x coords={32, 128, 512, 1024, 4096, 8192},
        xtick=data,
        nodes near coords,
        nodes near coords style={rotate=90, anchor=west, /pgf/number format/.cd,fixed zerofill,precision=2},
        ymin=70,ymax=100,
        ylabel={Accuracy},
    ]

    \addplot [style = {fill=color0}] table[x=images,y=lr]{\lr};	 
    \addplot [style = {fill=color1, mark=none, postaction={pattern=north east lines}}] table[x=images,y=rank-1-blr]{\blr};
    \addplot [style = {fill=color2, mark=none, postaction={pattern=north east lines}}] table[x=images,y=rank-2-blr]{\blr};
    \addplot [style = {fill=color3, mark=none, postaction={pattern=north east lines}}] table[x=images,y=rank-3-blr]{\blr};
    \addplot [style = {fill=color4, mark=none, postaction={pattern=north east lines}}] table[x=images,y=rank-4-blr]{\blr};
    
    \legend{LLR, Rank-$1$ BLR, Rank-$2$ BLR, Rank-$3$ BLR, Rank-$4$ BLR}
    \end{axis}
    \end{tikzpicture}
\caption{Comparaison of the accuracy of \gls{llr} and rank-$L$ \gls{blr} while performing classification of digits 5 and 8 with different training set size $T$ on \gls{mnist} data set.}
\label{figure: compare 58}
\end{figure}

\end{experiment}

\section{Generalisation to multiclass problem}
\label{section: Generalisation to multiclass classification problem}

The generalization of linear logistic regression that can perform multiclass classification is obtained via \gls{sr}: given $K$ classes, we have to calculate the posterior probability for each class $C_t\in\{1,2,\ldots,K \}$, \ie $y_{t,k}=\Pr\{C_t=k|\mat{X}_t\}, k=1,\ldots, K$. This is done using the following model:
\begin{equation}
\label{model: bsr}
y_{t,k}=
\frac{\exp\big(z_{t,k}\big)} {\sum_{j=1}^{K} \exp\big(z_{t,j}\big)},
\end{equation}
where 
\begin{align}\label{zk}
    z_{t,k}=\langle \mat{W}_k, \mat{X}_{t}\rangle,
\end{align}
and each $\mat{W}_k, k=1,\ldots, K$ represents the weights corresponding to the class $k$. For $K=2$, the indexing of the outputs with $k$ may be avoided as we did when discussing binary classification with \gls{llr}.

Using the same arguments as before, we will replace the inner product \eqref{zk} with  its bilinear counterpart:
\begin{align}
\label{zk.BL}
    z_{t,k}=\sum_{l=1}^L  \vec{a}_{l,k}^\T \mat{X}_{t} \vec{b}_{l,k},
\end{align}
which yields \gls{bsr}.

\subsection{Model training}
Given a training set consisting of pairs $\{(\mat{X}_t, \vec{c}_t)\}_{t=1}^{T}$ where $\vec{c}_t \in \mathbb{R}^{K}$ is the class-encoding vector such that $c_{t,k}=1$ if $C_{t}=k$, we want
to learn the weights $\mat{A}_l=[
\vec{a}_{l,1} , \cdots, \vec{a}_{l,K}]  \in \mathbb{R}^{M  \times K} $ and 
$ \mat{B}_l=[
\vec{b}_{l,1} , \cdots, \vec{b}_{l,K}] \in \mathbb{R}^{N \times K} $ for $l=1,\cdots,L$.

This is done by minimizing the following cost function 
\begin{equation}
J(\mat{A}, \mat{B}) 
= - \frac{1}{T}\sum_{t=1}^{T} \sum_{k=1}^{K} c_{t,k} \ln{y_{t,k}} 
+ \alpha R(\mat{A}, \mat{B}),
\label{cost: multiclass cross entropy bsr}
\end{equation}
where $\mat{A}=[\mat{A}_1,\ldots,\mat{A}_L]$ and $\mat{B}=[\mat{B}_1,\ldots,\mat{B}_L]$ gather the weights.

The gradient of the cost function $ J(\mat{A}, \mat{B}) $ with respect to each $ \vec{a}_{l,k} $ and $ \vec{b}_{l,k} $ is given by 

\begin{dmath}
\label{cost gradient: bsr with repect z}
\nabla_{\vec{a}_{l,k}} J(\mat{A}, \mat{B}) = \frac{1}{T} \sum_{t=1}^{T} \big( y_{t,k} - c_{t,k} \big) \mat{X}_t \vec{b}_{l,k} + \alpha \nabla_{\vec{a}_{l,k}} R(\mat{A}, \mat{B}),
\end{dmath}
\begin{dmath}
\label{cost gradient: bsr with respect to b}
\nabla_{\vec{b}_{l,k}} J(\mat{A}, \mat{B}) = \frac{1}{T} \sum_{t=1}^{T} \big( y_{t,k} - c_{t,k} \big) \vec{a}_{l,k}^\T \mat{X}_t  + \alpha \nabla_{\vec{b}_{l,k}} R(\mat{A}, \mat{B}).
\end{dmath}

We optimize $ J(\mat{A}, \mat{B}) $ using gradient descent as described in Algorithm~\ref{minimize cost: bsr} which generalizes \gls{blr} training defined in Algorithm~\ref{Algo:BLR} to the \gls{bsr} training.

\begin{algorithm}
	\SetAlgoLined
	\caption{Training of \gls{bsr}}
	\label{minimize cost: bsr}
	\text{Initialization:}\\
	$\mat{A} = \vec{0}$\\
	$\mat{B} = \text{drawn from a uniform distribution over [-1,1]}$\\
	\For{$k = 1, \cdots, K$}{
		\For{$l = 2, \cdots, L$}{
			$\mat{S}_{l-1,k } = \Big[ \frac{\vec{b}_{1,k}}{\norm{\vec{b}_{1,k}}}, \cdots, \frac{\vec{b}_{l-1,k}}{\norm{\vec{b}_{l-1,k}} }\Big] $
			
			$\vec{b}_{l,k} \leftarrow \vec{b}_{l,k} - \mat{S}_{l-1,k}\big(\mat{S}_{l-1,k}^\T \vec{b}_{l,k}\big)$
		}
	}
	
	\text{Optimization:}\\
	\For{$i = 1, \cdots, i_{\textnormal{max}}$}{
		\For{$l = 1, \cdots, L$}{
			\While{ $\mat{A}_l$ \textnormal{not converged} }{
    			$ \mat{G}_l = [\nabla_{\vec{a}_{l,1}} J(\mat{A}, \mat{B}), \cdots, \nabla_{\vec{a}_{l,K}} J(\mat{A}, \mat{B}) ]$
			
				$ \mat{D}_l = [ \mat{0}, \cdots, \mat{G}_l, \cdots, \mat{0} ]$ 
				
				$\hat{\eta} \approx \argmin_{\eta}{ J( \mat{A}-\eta\mat{D}_l, \mat{B} ) }$
				
				$\mat{A}_l \leftarrow \mat{A}_l - \hat{\eta}\mat{G}$
				
			}
			\While{ $\mat{B}_l$ \textnormal{not converged} }{
			
    			$ \mat{G}_l = [\nabla_{\vec{b}_{l,1}} J(\mat{A}, \mat{B}), \cdots, \nabla_{\vec{b}_{l,K}} J(\mat{A}, \mat{B}) ]$
    			
    			$ \mat{D}_l = [ \mat{0}, \cdots, \mat{G}_l, \cdots, \mat{0} ]$ 
				
				$\hat{\eta} \approx \argmin_{\eta}{ J( \mat{A}, \mat{B}-\eta\mat{D}_l ) }$
				
				$\mat{B}_l \leftarrow \mat{B}_l - \hat{\eta}\mat{G}_l$
				
			}
			
		}
	}
\end{algorithm}

\begin{experiment}
The \gls{bsr}  algorithm was applied to the same MNIST data set as in Experiment~\ref{Experiment.1v1} but using all $K=10$ classes. The validation set and the testing set contain $10000$ images. 

The classification accuracy is shown in Fig.~\ref{figure: lsr} and the conclusions are in line with those we drew in Experiment~\ref{Experiment.1v1}. The main difference is that the rank of the bilinear representation must be increased up to $L=4$ to obtain the results comparable with those yield by the linear soft-max regression (LSR). Thus, instead of $784K$ weights required in \gls{sr}, \gls{bsr} needs $224 K$ weights. 

\pgfplotstableread[row sep=\\,col sep=&]{
images & LSR  \\
160     & 78.14 \\
640    & 85.23\\
2560     & 88.88\\
5120	& 90.3   \\
}\lsr
\pgfplotstableread[row sep=\\,col sep=&]{
images  & rank-1-bsr	& rank-2-bsr    	& rank-3-bsr  \\
160      & 74.28        	& 78.2     			& 79.09  \\
640      & 81.76           	& 85.63      		& 85.64  \\
2560     & 84.53          	& 88.73      		& 89.19  \\
5120     & 85.34          	& 89.21      		& 89.94   \\
}\bsr
\begin{figure}
\centering
    \begin{tikzpicture}[font=\scriptsize]
    \begin{axis}[
       	ybar,
        bar width=5pt,
        x= 1.5cm,
        ybar=1pt,
        width=2\textwidth,
        height=.3\textwidth,
        hide y axis,
        axis x line*=bottom,
        legend style={at={(0.5,-0.2)}, anchor=north,legend columns=-1, column sep=0.1cm},
        symbolic x coords={160, 640, 2560, 5120},
        xtick=data,
        nodes near coords,
        nodes near coords style={rotate=90, anchor=west, /pgf/number format/.cd,fixed zerofill,precision=2},
        ymin=70,ymax=100,
        ylabel={Accuracy},
    ]

    \addplot [style = {fill=color0}] table[x=images,y=LSR]{\lsr};	 
    \addplot [style = {fill=color1, mark=none, postaction={pattern=north west lines}}] table[x=images,y=rank-1-bsr]{\bsr};
    \addplot [style = {fill=color2, mark=none, postaction={pattern=north west lines}}] table[x=images,y=rank-2-bsr]{\bsr};
    \addplot [style = {fill=color3, mark=none, postaction={pattern=north west lines}}] table[x=images,y=rank-3-bsr]{\bsr};
    
    \legend{LSR, Rank-$1$ BSR, Rank-$2$ BSR, Rank-$3$ BSR}
    \end{axis}
    \end{tikzpicture}
\caption{Comparaison of the accuracy of LSR and Rank-$L$ BSR while performing multiclass classification with different training set size $T$ on \gls{mnist} data set.}
\label{figure: lsr}
\end{figure}
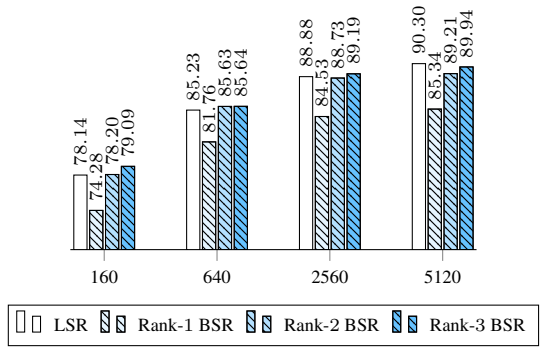
\end{experiment}

\section{Conclusions}\label{Sec:Discussion}
In this work we introduced and anaylzed the bilinear model to replace the linear operation used conventionally in the logistic regression. We also proposed a suitable optimization algorithm which exploits the convexity of the solution space; this allows us to obtain unique solution using gradient-based methods.

The solution was tested using MNIST data set of monocromatic images. We have shown that \gls{blr} can provide the same---and in some cases, better---performance as the conventional \gls{llr} which ignores the structure of the image. The results obtained using \gls{blr} require much less parameters which indicates that the overparametrization in the \gls{llr} is due to ignoring the correlation between the neighbourhood pixels.

{\small
\bibliographystyle{plain}
\bibliography{BMML_2019}
}

\end{document}